\pdfoutput=1

\documentclass[11pt]{article}

\usepackage{emnlp2021}

\usepackage{times}
\usepackage{latexsym}

\usepackage[T1]{fontenc}

\usepackage[utf8]{inputenc}

\usepackage{microtype}

\usepackage{url}
\usepackage{graphicx}
\usepackage{booktabs}
\usepackage{amssymb}
\usepackage{amsfonts}
\usepackage{multirow}
\usepackage{amsmath}

\usepackage{xspace,mfirstuc,tabulary}

\usepackage{color}

\DeclareMathOperator*{\argmin}{argmin}
\DeclareMathOperator*{\argmax}{argmax}

\title{Guilt by Association: Emotion Intensities in Lexical Representations}

\author{Shahab Raji \\
  Rutgers University \\
  \texttt{shahab.raji@rutgers.edu} \\\And
  Gerard de Melo \\
  Hasso Plattner Institute/University of Potsdam\\
  \texttt{gdm@demelo.org} \\}

\date{}

\begin{document}

\maketitle
\begin{abstract}
What do word vector representations reveal about the emotions associated with words? In this study, we consider the task of estimating word-level emotion intensity scores for specific emotions, exploring unsupervised, supervised, and finally a self-supervised method of extracting emotional associations from word vector representations.
Overall, we find that word vectors carry substantial potential for inducing fine-grained emotion intensity scores, showing a far higher correlation with human ground truth ratings than achieved by state-of-the-art emotion lexicons. 
\end{abstract}

\section{Introduction}
Vector representations of words induced using distributional co-occurrence signals provide valuable information about associations between words
\cite{baroni-etal-2014-dont}.
In this paper, we consider the question:
What do word vector representations reveal about the emotions associated with words?

There has been substantial research on methods to label words with associated emotions. 
Crowd-sourcing approaches have been used to compile databases to study the nexus between them.
\newcite{emolex} labelled around 15,000 uni\-grams with 8 basic emotions and two sentiments, providing binary tags for each word.
\newcite{nrc-word-affint} instead solicited real-valued emotion intensity ratings in $[0,1]$ for roughly 500--1800 words for eight emotions.
DepecheMood \cite{dm2014} and the improved DepecheMood++ \cite{dm++} are based on simple statistical methods to create an emotion lexicon from emotionally tagged data crawled from specific Web sites. 
We conjecture that vector representations allow us to obtain higher-quality ratings of the emotional associations of words.

Word vectors \cite{google-w2v,glove} have often been evaluated on standard word relatedness benchmarks. In this paper, we instead explore to what extent they encode emotional associations.
Earlier methods \cite{strapparava2008learning,mac2010evaluation} used corpus statistics in tandem with dimensionality reduction techniques for emotion prediction.
\newcite{rothe-etal-2016-ultradense} proposed predicting sentiment polarity ratings from word vectors, while other studies \cite{buechel-hahn-2018-word,Buechel2018Arxiv} predicted valence, arousal, and dominance using supervised deep neural networks. \newcite{Khosla2018Aff2Vec} proposed incorporating valence, arousal, and dominance ratings as additional signals into dense word embeddings. \newcite{buechel-hahn-2018-emotion} showed that emotion intensity scores can be predicted based on a lexicon providing valence--arousal--dominance ratings. 
We show that we can make use of word vectors to obtain high correlations with emotion intensities without any need for such ratings.

\paragraph{Contribution.}
Overall, our intriguing finding in this paper is that dense word vectors allow us to predict much more accurate emotion associations than state-of-the-art emotion lexicons.
We show that this holds even without any supervision, while different kinds of supervised setups yield even better results.

\section{Emotion Intensities as Associations}
\label{sec:groundtruth}

While many emotion lexicons only provide binary emotion labels, 
it is clear that words may exhibit different degrees of association with an emotion. 
For instance, the word \emph{party} suggests a greater degree of \emph{joy} than the word \emph{sea}, although the latter may as well be associated with joy.

\paragraph{Ground truth.}
The NRC Emotion/Affect Intensity Lexicon \cite{nrc-word-affint} provides human ratings of emotion intensity for words, scaled to $[0,1]$ such that a score of 1 signifies that a word ``conveys the highest amount of'' a particular emotion, while $0$ corresponds to the lowest amount. The real-valued scores were computed based on best--worst scaling of ratings solicited via crowdsourcing.

In our work, instead of viewing this resource as a lexicon that provides emotion intensity scores, we propose to treat it as providing \emph{associations between pairs of words}, one of the two words being an emotion. Thus, we propose to derive semantic association benchmarks similar to widely used semantic relatedness benchmark sets such as the RG-65 \cite{Rubenstein:1965:CCS:365628.365657} and WS353 \cite{Finkelstein:2001:PSC:371920.372094} test collections. 

\paragraph{Data split.}
In order to more fairly evaluate resources that only provide unigrams, we disregard bigrams in the lexicon,
resulting in a total of 9,706 word pairs. The data covers the eight basic emotions proposed in Plutchik's wheel of emotions \cite{plutchik1980}, i.e.,
\emph{anger}, \emph{anticipation}, \emph{disgust}, \emph{fear}, \emph{joy}, \emph{sadness}, \emph{surprise}, and \emph{trust}.
To facilitate an evaluation of unsupervised and supervised techniques, for each emotion, we split the corresponding data into training/validation/test portions with a ratio of 64\%/16\%/20\%. This results in a total of 7,762 pairs in the training sets, and 1,944 pairs in the test sets. The remaining 1,552 instances serve as validation data.

\section{Estimating Emotion Intensity}
\label{sec:methods}
Emotion lexicons are typically created using crowdsourcing \cite{emolex} or  drawing on emotion-labeled text \cite{dm2014,dm++}.
We investigate deriving emotion scores from word vector representations without any emotion-specific signals.
\begin{table*}[htbp]
\centering
\resizebox{\linewidth}{!}{
\small
\begin{tabular}{lllrrrrrrrrr}
\toprule
\multicolumn{2}{l}{\textbf{Method}} & \textbf{Data}& \textbf{Anger} & \textbf{Anticipation} & \textbf{Disgust} & \textbf{Fear} & \textbf{Joy} & \textbf{Sadness} & \textbf{Surprise} & \textbf{Trust} & \textbf{Overall} \\	 
\midrule

{\multirow{4}{*}{\rotatebox[origin=c]{90}{Baselines}}} &
& EmoLex        &
0.081 & \textbf{-0.071} & \textbf{-0.037} & 0.101 & \textbf{0.059} & 0.035 & \textbf{-0.186} & \textbf{0.250} & 0.076 \\
&& DepecheMood$^{*}$   &
0.089 &-- &--& 0.077 & -0.020 & 0.204 &--&--& 0.042 \\
&& DepecheMood++$^{*}$ &
\textbf{0.207} &--&--& \textbf{0.180} & 0.010 & \textbf{0.337} &--&--& \textbf{0.106} \\
&& EmoWordNet$^{*}$    &
0.100 &--&--& 0.077 & 0.032 & 0.190 &--&--& 0.048 \\

\midrule
\midrule

{\multirow{7}{*}{\rotatebox[origin=c]{90}{Unsupervised}}}&
& word2vec &
0.321 & 0.266 & 0.056 & 0.081 & 0.482 & 0.321 & 0.249 & 0.403 & 0.254\\

&& GloVe &
0.360 & 0.413 & 0.031 & 0.137 & 0.352 & 0.281 & -0.025 & 0.439 & 0.246\\

&& fastText &
0.388 & 0.332 & 0.071 & 0.038 & 0.306 & 0.281 & 0.186 & 0.497 & 0.249\\

&& EWE &
0.301 &  \textbf{0.434} &  0.120 &  0.127 &  0.236 &  0.181 &  -0.143 & 0.348 & 0.194\\

&& Counter-fitting &
0.354 & 0.379 & 0.075 & 0.200 & 0.507 & 0.379 & \textbf{0.334} & 0.459 & 0.322\\

&& AffectVec &
\textbf{0.426} & 0.430 & \textbf{0.262} & \textbf{0.447} & \textbf{0.631} & \textbf{0.559} & -0.078 & \textbf{0.592} & \textbf{0.419}\\
&& BERT base &
0.130 & 0.104 & -0.136 & 0.081 & 0.002 & -0.012 & -0.037 & 0.155 & 0.033\\

\midrule
\midrule

{\multirow{7}{*}{\rotatebox[origin=c]{90}{Self-sup.}}}&
{\multirow{7}{*}{\rotatebox[origin=c]{90}{FFNN}}}

 & word2vec &
	 0.339   & 0.229   & 0.300   & 0.215   & 0.412   & 0.447   & 0.103   & 0.311   &0.311\\

&& GloVe &
	 0.266   & 0.256   & 0.214   & 0.158   & 0.351   & 0.386   & 0.024   & 0.207   &0.239\\

&& fastText &
	 0.352   & 0.300   & \textbf{0.348}   & 0.272   & 0.347   & 0.530   & \textbf{0.137}   & 0.469   &0.359\\
&& EWE &
	 0.259   & 0.242   & 0.251   & 0.174   & 0.330   & 0.304   &-0.086   & 0.270   & 0.228\\

&& Counter-fitting &
	 0.345   & 0.305   & 0.177   & 0.205   & 0.467   & 0.353   & 0.119   & 0.445   &0.300\\
&& AffectVec &
	 \textbf{0.411}   & \textbf{0.418}   & 0.273   & \textbf{0.417}   & \textbf{0.631}   & \textbf{0.564}   &-0.091   & \textbf{0.574}   &\textbf{0.405}\\
&& BERT base &
    0.347   & 0.081   & 0.259   & 0.080   & 0.389   & 0.327   & 0.114   & 0.310   & 0.238\\

\midrule

{\multirow{7}{*}{\rotatebox[origin=c]{90}{Self-sup.}}}&
{\multirow{7}{*}{\rotatebox[origin=c]{90}{SVR}}}
 & word2vec &
	 0.410   & 0.227   & 0.311   & 0.226   & 0.331   & 0.472   & 0.098   & 0.361   &0.303\\

&& GloVe &
	 0.419   & 0.379   & 0.335   & 0.256   & 0.494   & 0.509   & 0.061   & 0.441   &0.386\\

&& fastText &
	 0.346   & 0.302   & \textbf{0.409}   & 0.272   & 0.389   & 0.513   & \textbf{0.158}   & 0.480   &0.371\\
	 
&& EWE &
	 0.348   & 0.327   & 0.386   & 0.249   & 0.390   & 0.486   &-0.094   & 0.487   & 0.322\\

&& Counter-fitting &
	 0.368   & 0.349   & 0.224   & 0.213   & 0.471   & 0.388   & 0.138   & 0.469   &0.314\\

&& AffectVec &
	 \textbf{0.437}   & \textbf{0.457}   & 0.302   & \textbf{0.490}   & \textbf{0.633}   & \textbf{0.600}   &-0.128   & \textbf{0.605}  &\textbf{0.426}\\
&& BERT base &
	 0.331   & 0.029   & 0.287   & 0.029   & 0.374   & 0.356   & 0.079   & 0.309   & 0.248\\

\midrule
\midrule

\multirow{7}{*}{\rotatebox[origin=c]{90}{Supervised}}&
\multirow{7}{*}{\rotatebox[origin=c]{90}{FFNN}}
 & word2vec &
	 0.681   & 0.568   & 0.711   & 0.713   & 0.695   & \textbf{0.749}   & 0.712   & 0.694   &0.699\\
&& GloVe &
	 0.704   & 0.559   & 0.733   & 0.723   & 0.678   & 0.731   & 0.643   & 0.730   &0.698\\
&& fastText &
	 \textbf{0.716}   & 0.596   & \textbf{0.761}   & \textbf{0.736}   & 0.691   & 0.743   & \textbf{0.756}   & \textbf{0.750}   &\textbf{0.722}\\
&& EWE &
	 0.582   & 0.322   & 0.618   & 0.636   & 0.582   & 0.662   & 0.438   & 0.602   & 0.554\\
&& Counter-fitting &
	 0.675   & 0.580   & 0.641   & 0.676   & 0.660   & 0.649   & 0.732   & 0.693   &0.665\\
&& AffectVec &
	 0.700   & \textbf{0.614}   & 0.664   & 0.703   & \textbf{0.724}   & 0.686   & 0.747   & 0.701   &0.695\\
&& BERT base &
	 0.624   & 0.486   & 0.550   & 0.578   & 0.578   & 0.570   & 0.612   & 0.580   & 0.533\\

\midrule

{\multirow{7}{*}{\rotatebox[origin=c]{90}{Supervised}}}&
{\multirow{7}{*}{\rotatebox[origin=c]{90}{SVR}}}
 & word2vec &
	 0.642   & 0.566   & 0.654   & 0.654   & 0.680   & \textbf{0.734}   & 0.721   & 0.664   &0.672\\
&& GloVe &
	 \textbf{0.720}   & \textbf{0.633}   & 0.727   & 0.671   &\textbf{0.719}   & 0.723   & 0.722   & 0.694   &\textbf{0.705}\\
&& fastText &
	 0.660   & 0.589   & \textbf{0.745}   & \textbf{0.698}   & 0.637   & 0.697   & \textbf{0.746}   & 0.700   &0.685\\
	 
&& EWE &
	 0.685   & 0.618   & 0.705   & 0.633   & 0.642   & 0.720   & 0.709   & \textbf{0.713}   & 0.677\\
&& Counter-fitting &
	 0.654   & 0.569   & 0.558   & 0.623   & 0.618   & 0.636   & 0.711   & 0.665   &0.630\\
&& AffectVec &
	 0.678   & 0.606   & 0.631   & 0.676   & 0.657   & 0.671   & 0.668   & 0.675   &0.664\\
&& BERT base &
	 0.567   & 0.478   & 0.484   & 0.532   & 0.576   & 0.559   & 0.683   & 0.556   & 0.553\\

\bottomrule
\end{tabular}
}
\caption{Main results for emotion intensity prediction, reported in terms of Pearson Correlation. ($^{*}$ These baselines are averaged based only on the labels they cover.)}
\label{tab:main8-results}
\end{table*}

\subsection{Unsupervised Prediction}
\label{unsupervised_method}

While past work on emotion intensity prediction has considered this an entirely separate task, we here consider emotional intensity scoring as similar in nature to regular lexical associations between words. 
Given two words $w_1$, $w_2$, the cosine similarity 
$\mathrm{cos}(\mathbf{v}_{w_1}, \mathbf{v}_{w_2})$ 
of their corresponding word vectors $\mathbf{v}_{w_1}$, $\mathbf{v}_{w_2}$ is expected to reflect their association, as most word vector spaces capture semantic relatedness \cite{simlex}.

Hence, for an unsupervised estimation of emotion intensities, 
we simply look up the association between a targeted vocabulary word $w$ and the emotion $e$ under consideration, obtaining
\begin{equation}
\sigma_\mathrm{u}(w,e) = \mathrm{cos}(\mathbf{v}_{w}, \mathbf{v}_{w_e}) = \frac{\mathbf{v}_{w} \cdot \mathbf{v}_{w_e}}{|\mathbf{v}_{w}|\,|\mathbf{v}_{w_e}|}\label{eq:unsupervised}
\end{equation}
as the emotion intensity score. Here, we assume there is
a single word $w_e$ denoting $e$ (e.g., \emph{joy}).

\subsection{Supervised Prediction}\label{sec:supervised}

If a training set is available,
then for each emotion $e$ covered by it, we train a regression model $f_e(\mathbf{v}_{w} \mid \theta_e)$ parametrized by $\theta_e$ that allows us to predict the emotion intensity for $e$, given a vector of a word as input. Finally, we simply define
\begin{equation}
\sigma_\mathrm{s}(w,e) = f_e(\mathbf{v}_{w} \mid \theta_e)\label{eq:supervised}
\end{equation}
to invoke the
relevant $f_e$ based on the emotion $e$ under consideration.

We consider two kinds of models. The first is a feedforward neural network with a ReLU activated hidden layer with 64 neurons and a single output neuron that predicts the intensity of emotion, and a dropout rate of 0.2 applied to the hidden layer. The loss function is the mean squared error of the prediction with respect to the ground truth.
The second option is a support vector regression (SVR) model with RBF kernel.

\subsection{Self-Supervised Prediction}
Finally, we propose a hybrid self-supervised technique that relies on supervised learning but does \textbf{\emph{not}} require any pre-existing training data. Instead, for each emotion $e$, we first identify sets
\begin{align}
    T_e^{+}=\argmax\limits_{T \subset \mathcal{V}, |T|=k}\,\sum_{w\in T} \sigma_\mathrm{u}(w,e)\\
    T_e^{-}=\argmin\limits_{T \subset \mathcal{V}, |T|=k}\,\sum_{w\in T} \sigma_\mathrm{u}(w,e)
\end{align}
of words in the vocabulary $\mathcal{V}$ with the top $k$ highest and lowest intensity predictions
as defined by Eq.~\ref{eq:unsupervised}, i.e., our unsupervised 
technique.
For each $e$, $T_e^{+} \cup T_e^{-}$ along with the corresponding labels is used to train a supervised model $\bar{f}_e(\mathbf{v}_{w} \mid \bar{\theta}_e)$ as above in Section~\ref{sec:supervised}. Finally, the overall model again just involves
invoking the relevant
$\bar{f}_e$:
\vspace*{-0.2cm}
\begin{equation}
\sigma_\mathrm{\bar{s}}(w,e) = \bar{f}_e(\mathbf{v}_{w} \mid \bar{\theta}_e)\label{eq:selfsupervised}
\end{equation}

\section{Evaluation}\label{sec:eval}

\subsection{Main Results}

We evaluate to what extent word vectors capture emotion intensity information by comparing the presented methods 
against the ground truth test sets from Section \ref{sec:groundtruth}.
The score for out-of-vocabulary words is taken to be 0.0 and we retain the same test set for each emotion.

Table \ref{tab:main8-results} provides an overview of the results for a series of baselines along with the unsupervised, self-supervised, and regular supervised methods proposed in Section \ref{sec:methods}. 
The ``Overall'' column reports the correlation with the union of all word--emotion pairings in the ground truth test set. As we have an equal number of word--emotion pairs for each emotion, this serves as an aggregate measure of the result quality.

\paragraph{Baselines.} 
We first evaluate the state-of-the-art emotion lexicons,
finding that the scores that they provide exhibit very low correlation with the ground truth. In the case of EmoLex \cite{emolex}, this is because it merely provides binary labels, not intensities. 

For DepecheMood \cite{dm2014}, DepecheMood++ \cite{dm++}, and EmoWordNet \cite{badaroetal2018emowordnet}, we conjecture that the data-driven automated techniques used to create them based on coarse-grained document-level labels do not result in word-level scores of the same sort as those solicited from human raters. The additional labels in DepecheMood and DepecheMood++ (e.g., Amused) may carry some information on some of the labels in our ground truth data (e.g., Joy). However, since the term--document matrices are calculated independent of each other, mapping these emotions to a target emotion will result in less accurate emotion association scores.

\paragraph{Unsupervised method.} 
We next evaluate various pretrained word vector models against the test set, including
pretrained word2vec trained on Google News \cite{google-w2v}, 
Glove 840B CommonCrawl embeddings \cite{glove}, and fastText trained on Wikipedia \cite{fasttext-joulin-etal-2017}. Among these, we find that fastText vectors outperform word2vec and GloVe. They also outperform the pretrained BERT-base (uncased) model \cite{devlin2019bert}, for which we use the average of word-piece-level output embeddings as word-level vectors.
Even higher correlation can be attained with vectors that have been post-processed.
One example are the counter-fitted vectors \cite{counterfitting-MrksicSTGRSVWY16}, obtained by taking the PARAGRAM-SL999 vectors by \newcite{WietingEtAl2015} and optimizing them using synonymy and antonymy constraints. 
Better results are obtained by the AffectVec technique \cite{RajiDeMelo2020AffectVec}, which post-processes the same vectors using not only synonymy and antonymy constraints, but also sentiment polarity ones. Sentiment polarity evidently helps to better distinguish different emotional associations. In contrast, the emotion-enriched word vectors (EWE) introduced by \newcite{agrawal-etal-2018-learning} do not perform well for word intensity prediction.

\paragraph{Supervised method.} 

For supervised methods, we report the mean correlation over 20 runs of the learning algorithm.
As expected, the models succeed in learning emotional associations from word vectors with a higher correlation than unsupervised techniques. As this technique does not rely on cosine similarities, post-processing word vector spaces proves fruitless, and very strong results are obtained using GloVe and fastText embeddings.

\paragraph{Self-supervised method.}
Unsupervised prediction is less accurate than supervised prediction, but is applicable to arbitrary emotions without any need for pre-existing emotion-specific training data. The self-supervised approach has the same advantage of not requiring training data and can thus as well be applied to arbitrary emotions.
We evaluate it for $k=100$, i.e., with 200 automatically induced training instances per emotion. For BERT vectors, we use the fastText vocabulary to find the most and least similar words to the emotion. As expected, without access to gold standard training data, self-supervision is unable to compete with the supervised approach. However, it is able to surpass the unsupervised approach despite drawing on it for training, owing to its ability to selectively pick out the most pertinent cues from the word vector.

\subsection{Unsupervised Sentence Classification}
Finally, we explore using our results for unsupervised sentence-level emotion classification on the GoEmotions dataset \cite{Demszky2020} with its fine-grained inventory of 28 different labels. 
Emotion lexicons are often used for unsupervised analysis, but existing lexicons do not cover all emotions in this dataset (e.g., \emph{remorse}, \emph{gratitude}, \emph{caring}).
Given input document $D$, we choose the label
\begin{equation}
\underset{e\in \Sigma}{\operatorname{arg\,max}}\, \sum_{w \in D}\gamma_{w,D} \,\sigma_u(e,w),\end{equation}
where $\Sigma$ is the set of labels, $\gamma_{w,D}$ denotes the TF-IDF score of $w$ in $D$, and $\sigma_u(w,e)$ is our unsupervised word scoring (Eq.~\ref{eq:unsupervised}).
An exception is made for the \emph{neutral} label, which we choose if the average prediction score across labels in $\Sigma$ is $\leq 0.03$,
based on the reasoning that, similar to the annotation instructions in the original paper, low scores or conflicting ones should be labeled neutral.

The results in Table \ref{tab:go-emotions} show that an entirely unsupervised approach is able to greatly outperform the random baseline. For reference, we also show the results of the \emph{supervised} BERT model from \newcite{Demszky2020}, while our models are entirely \emph{unsupervised}.
Note that this dataset is fairly challenging, since many of the 28 emotions are easy to confuse, e.g., \emph{fear}, \emph{nervousness}, \emph{embarrassment}, \emph{disappointment}, \emph{disapproval}, etc.

\begin{table}[]
\centering
\resizebox{0.8\linewidth}{!}{
\small
\begin{tabular}{lccc}
\toprule
\textbf{Method} & \textbf{Precision} & \textbf{Recall} & \textbf{F$_1$}\\
\midrule
Random          & \phantom{0}3.6 & \phantom{0}3.6 & \phantom{0}3.6 \\
\midrule
\textbf{Unsupervised}\\
~~~word2vec        & 26.2 & 13.7 & 11.2 \\
~~~GloVe           & 30.0 & \phantom{0}8.0 & \phantom{0}2.3 \\
~~~fastText        & 30.1 & 10.9 & \phantom{0}5.8 \\
~~~Counter-fitting  & 23.7 & 20.3 & 16.8 \\
~~~AffectVec       & 25.7 & 20.0 & 16.8 \\
\midrule
\textbf{Supervised}\\
~~~BERT & 40\phantom{0} & 63\phantom{0} & 46\phantom{0}\\
\bottomrule
\end{tabular}}
\caption{Results for 28-way unsupervised sentence classification on GoEmotions dataset, in terms of macro-averaged precision, recall, and F$_1$-score in \%.}
\label{tab:go-emotions}
\end{table}

\section{Conclusion}

In this paper, we 
show that pretrained word vectors readily provide higher-quality emotional intensity information
than state-of-the-art emotion lexicons, 
despite not being trained on emotion-labeled data, particularly if the word vector space is post-processed.
We further investigate supervised learning and
find that a regular supervised variant obtains very high correlations, while
a self-supervised variant that does not require gold standard training data is able to outperform the unsupervised method and can as well be applied to arbitrary emotion labels.
Overall, our results confirm that word vectors have remained underexplored for word-level emotion intensity assessment. We will share our training/validation/test splits and lexicons to promote further research in this area.

\section*{Broader Impact} 
Assessing the emotions evoked by a particular text has important applications, such as discovering disappointed customers on social media, designing conversational agents that emulate human empathy by responding in a more appropriate way, as well as numerous forms of digital humanities analyses. However, there are also important concerns and risks when using automated techniques to assess the emotional impact of text. Clearly, consent to assess the text must have been granted and malicious as well as potentially harmful applications must be avoided. Beyond this, however, even for applications deemed beneficial, there is a risk that inaccurate assessments may lead to undesirable outcomes. The empirical results in this paper show that many commonly used resources and techniques have a low correlation with human ratings and even the best BERT-based learning approaches are highly imperfect, as they may latch on to superficial lexical cues and mere correlations that may exhibit substantial biases. This is discussed further by \newcite{mohammad2020practical}. As a result, prior to any potential use of such techniques, a thorough analysis of potential risks must be conducted.

\appendix

\end{document}